\PassOptionsToPackage{table}{xcolor}
\documentclass[letterpaper]{article} 
\usepackage{tgtermes}
\makeatletter
\@namedef{ver@newtxtext.sty}{2024/04/01}
\makeatother
\usepackage{aaai2027}  
\usepackage[hyphens]{url}  
\usepackage{graphicx} 
\usepackage{natbib}  
\usepackage{caption} 
\usepackage{booktabs}
\usepackage{pifont}
\usepackage{amsmath}
\usepackage{colortbl}
\usepackage{placeins}
\usepackage{pgf}

\makeatletter
\newcommand{\cell}[1]{%
  \pgfmathparse{max((#1-25)*45/75, 0)}%
  \xdef\@cellintensity{\pgfmathresult}%
  \cellcolor{green!\@cellintensity!white}#1}
\newcommand{\cellb}[1]{%
  \pgfmathparse{max((#1-25)*45/75, 0)}%
  \xdef\@cellintensity{\pgfmathresult}%
  \cellcolor{green!\@cellintensity!white}\textbf{#1}}
\newcommand{\cellu}[1]{%
  \pgfmathparse{max((#1-25)*45/75, 0)}%
  \xdef\@cellintensity{\pgfmathresult}%
  \cellcolor{green!\@cellintensity!white}\underline{#1}}
\makeatother

\newcommand{\cmark}{\ding{51}}
\newcommand{\dash}{--}

\title{Cyc3D: Evaluating Cyclic Structural Stability and Asset Usability in Image-to-3D Generation}
\author{
    Liwen Zhang
}
\affiliations{}

\begin{document}

\maketitle

\begin{abstract}
Image-conditioned 3D generation has advanced rapidly, yet existing evaluation protocols largely judge rendered-view plausibility and semantic alignment, overlooking whether a generator forms a stable 3D interpretation and produces assets usable in graphics pipelines. We introduce Cyc3D, a multidimensional benchmark that evaluates image-to-3D generation along two complementary axes: Cross-View Object Consistency and Representation Quality. At the asset level, Cyc3D measures whether object identity remains semantically coherent across rendered viewpoints. At the model level, we propose View-Cycle Structural Consistency, a closed-loop render--regenerate--align protocol that repeatedly re-observes a generated asset from novel views and quantifies geometric, perceptual, and semantic drift across generations. To assess native asset usability beyond rendered appearance, Cyc3D further evaluates geometric structure, reference-image fidelity, mesh discretization and efficiency, and UV parameterization quality. Together, these diagnostics expose failures obscured by a single perceptual score and provide interpretable evidence of both model instability and representation defects. Experiments on five representative image-to-3D systems show that closed-source feed-forward models consistently outperform open-source optimization-based baselines in geometric fidelity, mesh quality, and cycle stability. Nevertheless, even the strongest methods achieve cycle-stability scores below 48, revealing a persistent gap between visually plausible generation and robust 3D object understanding.
\end{abstract}

\section{Introduction}

High-quality 3D assets are fundamental to digital content creation, games, architectural design, robotics, and embodied simulation. With the rapid progress of generative models, generating a 3D asset from a single reference image has become an increasingly important capability in modern 3D generation. Compared with text-to-3D generation, image-to-3D generation provides stronger instance-level control: the input image specifies not only the semantic category of an object, but also its identity, shape, color, texture, and many local visual details. Recent image-conditioned systems have demonstrated the ability to lift a single image into a full 3D textured object by combining multi-view generation, 3D reconstruction, or 3D-native diffusion models~\cite{liu2024one}. Nevertheless, evaluating image-to-3D generation remains challenging. A generated result may look plausible from a few rendered views, while the underlying 3D hypothesis can still be unstable, incomplete, or unsuitable for downstream asset workflows. Therefore, evaluating image-to-3D models should not only ask whether the generated asset appears visually reasonable from several viewpoints, but also whether it reflects a reliable understanding of the object's 3D identity, structure, and surface representation.

We argue that such reliability should be studied from two complementary perspectives. The first is \emph{Cross-View Object Consistency}, which measures whether an object remains stable under changes of viewpoint. Human perception naturally maintains a coherent object-level interpretation across views: although the visible appearance changes as the viewpoint moves, the object's identity and basic 3D structure should not fundamentally change. Existing evaluation work has already recognized the importance of multi-view assessment for 3D generation. GPTEval3D evaluates generated 3D assets through multi-view RGB and normal renderings under criteria such as text--asset alignment, 3D plausibility, geometry details, texture details, and geometry--texture coherence~\cite{wu2024gpt}. Gen3DEval further trains a vision-language evaluator for text fidelity, appearance, and surface quality using rendered views and surface normals~\cite{maiti2025gen3deval}. Eval3D provides a more diagnostic protocol by decomposing generated-asset quality into geometric, structural, semantic, text-alignment, and aesthetic aspects, and evaluates consistency through interpretable probes across views~\cite{duggal2025eval3d}. These works show that 3D generation evaluation is moving beyond single-view appearance toward more 3D-aware and consistency-aware analysis.

The second perspective is \emph{Representation Quality}, which concerns the quality of the 3D representation itself. A generated object is not merely an image-like rendering; it is a 3D asset that should be rendered, edited, textured, converted, and reused in downstream pipelines. Existing benchmarks have begun to evaluate geometry, texture, and surface quality. For example, Gen3DEval considers surface quality in addition to text fidelity and appearance~\cite{maiti2025gen3deval}; MATE-3D collects multi-dimensional human annotations over semantic alignment, geometry quality, texture quality, and overall quality~\cite{zhang2025benchmarking}; and Eval3D studies geometric, structural, and semantic failure modes in a fine-grained manner~\cite{duggal2025eval3d}. However, many existing metrics still primarily operate on rendered images, videos, captions, normal maps, or human/VLM preferences. These signals are valuable for perceptual evaluation, but they do not fully characterize the native usability of a generated 3D asset. In practical content creation, an asset may appear acceptable in rendered views while still suffering from representation-level defects such as inaccurate geometric structure, inefficient or degenerate mesh tessellation, poor surface discretization, or low-quality UV parameterization. These issues may not always lead to obvious semantic drift, but they directly affect rendering efficiency, texture editing, material adjustment, asset conversion, and downstream reuse.

In this paper, we introduce an image-to-3D benchmark that evaluates the 3D object modeling ability of current generative models along the two axes of Cross-View Object Consistency and Representation Quality. For Cross-View Object Consistency, we consider both asset-level and model-level reliability. At the asset level, we evaluate whether a generated 3D asset preserves object identity and basic structural characteristics across rendered viewpoints. At the model level, we further ask whether the generator forms a stable and self-consistent 3D interpretation of the same object. To this end, we propose \emph{cyclic structural consistency}: starting from a generated 3D asset, we render it from a novel viewpoint, feed the rendered image back into the image-to-3D generator, and compare the point-cloud geometry of the reconstructed asset with that of the previous step. Repeating this process forms a closed-loop test that measures structural drift across generations. If a model has a robust 3D prior for object structure, its reconstructions should remain geometrically stable under this generate--render--regenerate cycle, rather than exhibiting identity drift, geometric deformation, or progressive detail collapse.

For Representation Quality, we complement perceptual and multi-view evaluation with explicit analysis of asset-level representations. In addition to reference-image fidelity, which measures whether the generated object preserves the identity and visual details specified by the input image, we introduce metrics for mesh quality and UV quality. Mesh quality reflects whether the generated surface is geometrically well-formed and suitable for further processing, while UV quality measures whether the 2D parameterization supports reliable texture mapping, editing, and rendering. By incorporating reference-image adherence, mesh validity, UV parameterization, and cyclic structural consistency into a unified benchmark, our evaluation extends prior asset-level protocols toward a broader assessment of both generated assets and the generative models behind them.

In summary, the main contributions of this paper are as follows:

\begin{itemize}
    \item We propose a comprehensive evaluation framework for image-to-3D generation, aiming to systematically diagnose the reliability of current models in 3D object modeling. The framework is organized around two complementary dimensions: \emph{Cross-View Object Consistency} and \emph{Representation Quality}. The former evaluates whether the generated object preserves a stable identity and basic structural characteristics under viewpoint changes, while the latter examines whether the resulting 3D asset provides a reliable, processable, and reusable underlying representation.

    \item We extend image-to-3D evaluation from the asset level to the model level. Beyond assessing multi-view semantic identity consistency within a generated asset, we introduce \emph{View-Cycle Structural Consistency}, a closed-loop evaluation protocol that follows a generate--novel-view render--regenerate process. This metric tests whether an image-to-3D pipeline can produce structurally compatible object interpretations after viewpoint changes, thereby revealing the stability of the model's underlying 3D object understanding.

    \item We incorporate representation-level asset quality into image-to-3D evaluation. In addition to geometric structure and reference-view consistency, we explicitly assess \emph{mesh discretization quality} and \emph{UV parameterization quality}. These metrics complement existing evaluation protocols by capturing the low-level usability of generated 3D assets, which is critical for rendering, editing, texturing, conversion, and downstream reuse.

    \item We conduct a systematic evaluation of representative image-to-3D models under the proposed framework. Our analysis reveals that closed-source feed-forward systems (Hunyuan3D, Tripo3D) achieve substantially higher mesh quality, geometric fidelity, and cycle stability than open-source optimization-based methods (Stable Zero123, Magic123, DreamCraft3D), while all evaluated methods still exhibit notable cyclic consistency degradation, indicating that robust 3D object understanding remains an open challenge even for the strongest current generators.
\end{itemize}

Table~\ref{tab:benchmark_comparison} positions Cyc3D against representative 3D generation benchmarks and highlights its coverage of native asset quality and model-level cyclic consistency.

\begin{table*}[t]
\centering
\small
\setlength{\tabcolsep}{4.5pt}
\renewcommand{\arraystretch}{1.15}
\caption{Comparison with representative 3D generation benchmarks. Most existing benchmarks focus on asset-level evaluation, such as visual quality, texture quality, and cross-view consistency. Cyc3D further covers reference-image adherence, mesh quality, UV quality, and model-level cyclic consistency.}
\label{tab:benchmark_comparison}
\resizebox{\linewidth}{!}{
\begin{tabular}{lccccccccc}
\toprule
\textbf{Benchmark}
& \multicolumn{7}{c}{\textbf{Asset-level Evaluation}}
& \multicolumn{2}{c}{\textbf{Model-level Evaluation}} \\
\cmidrule(lr){2-8} \cmidrule(lr){9-10}
& \multicolumn{3}{c}{\textbf{Quality}}
& \multicolumn{2}{c}{\textbf{Editability}}
& \multicolumn{2}{c}{\textbf{Cross-view Consistency}}
& \multicolumn{2}{c}{\textbf{Cyclic Consistency}} \\
\cmidrule(lr){2-4} \cmidrule(lr){5-6} \cmidrule(lr){7-8} \cmidrule(lr){9-10}
& Appearance
& Texture
& Ref. Adherence
& Mesh
& UV
& Semantic
& Structural
& Semantic
& Structural \\
\midrule
T3Bench
& \dash & \dash & \dash
& \dash & \dash
& \cmark & \dash
& \dash & \dash \\

GPT-4V(ision) / GPTEval3D
& \cmark & \cmark & \dash
& \dash & \dash
& \cmark & \cmark
& \dash & \dash \\

Gen3DEval
& \cmark & \cmark & \dash
& \dash & \dash
& \cmark & \cmark
& \dash & \dash \\

Hi3DEval
& \cmark & \cmark & \dash
& \dash & \dash
& \cmark & \cmark
& \dash & \dash \\

Eval3D
& \cmark & \cmark & \dash
& \dash & \dash
& \cmark & \cmark
& \dash & \dash \\

\midrule
\textbf{Cyc3D (ours)}
& \cmark & \cmark & \cmark
& \cmark & \cmark
& \cmark & \cmark
& \cmark & \cmark \\
\bottomrule
\end{tabular}
}
\vspace{2mm}
\begin{minipage}{0.98\linewidth}
\footnotesize
\textbf{Note.}
A check mark indicates that the benchmark explicitly evaluates the corresponding aspect.
``Ref. Adherence'' refers to fidelity to the input reference image, which is particularly important for image-to-3D generation.
``Mesh'' and ``UV'' denote explicit evaluation of native asset representations rather than indirect assessment through rendered views.
``Model-level cyclic consistency'' refers to evaluating whether the image-to-3D model maintains a stable 3D object hypothesis under a generate--render--regenerate cycle.
\end{minipage}
\end{table*}

\section{Towards a Reliable Image-to-3D Benchmark}
\label{sec:towards}

\subsection{Image-to-3D Generation as 3D Object Modeling}
\label{sec:problem_formulation}

Given a reference image $I$, an image-to-3D generator $G_{\theta}$ produces a 3D asset
\begin{equation}
A = G_{\theta}(I).
\end{equation}
We view this process not merely as translating an image into an output asset, but as a form of 3D object modeling from limited 2D observation. A single image provides only partial evidence of the object. To generate a complete 3D asset, the model must infer an underlying 3D object hypothesis that explains the object's identity, global shape, visible and invisible geometry, surface appearance, and texture layout.

Conceptually, we decompose this process as
\begin{equation}
h = H_{\theta}(I), \qquad A = E_{\theta}(h),
\end{equation}
where $h$ denotes the model's inferred 3D object hypothesis, and $E_{\theta}$ realizes this hypothesis as an explicit 3D asset. This decomposition is used only as an evaluation abstraction: the generator does not need to expose $H_{\theta}$ and $E_{\theta}$ as separate modules.

A generated 3D asset can be further abstracted as
\begin{equation}
A = (\mathcal{M}, \mathcal{U}, \mathcal{T}),
\end{equation}
where $\mathcal{M}$ denotes the geometric representation, $\mathcal{U}$ denotes the surface parameterization such as UV mapping, and $\mathcal{T}$ denotes texture or material maps. For a mesh-based asset, $\mathcal{M}=(V,F)$ consists of vertices $V$ and faces $F$, while $\mathcal{U}$ maps the 2D texture domain to the 3D surface. A rendered image is only one observation of the underlying asset:
\begin{equation}
x_v = R(A; v, \ell),
\end{equation}
where $R$ is a renderer, $v$ is the camera viewpoint, and $\ell$ denotes lighting and rendering conditions.

This formulation highlights why evaluating only rendered appearance is insufficient. A generated object may look plausible from selected viewpoints, while its inferred 3D hypothesis may be unstable or its explicit asset representation may be unsuitable for downstream use. Therefore, a reliable image-to-3D benchmark should evaluate both the stability of the model's 3D object understanding and the quality of the generated asset representation.

\subsection{What Makes Reliable 3D Object Modeling?}
\label{sec:reliable_modeling}

\begin{figure*}[t]
\centering
\includegraphics[width=\textwidth]{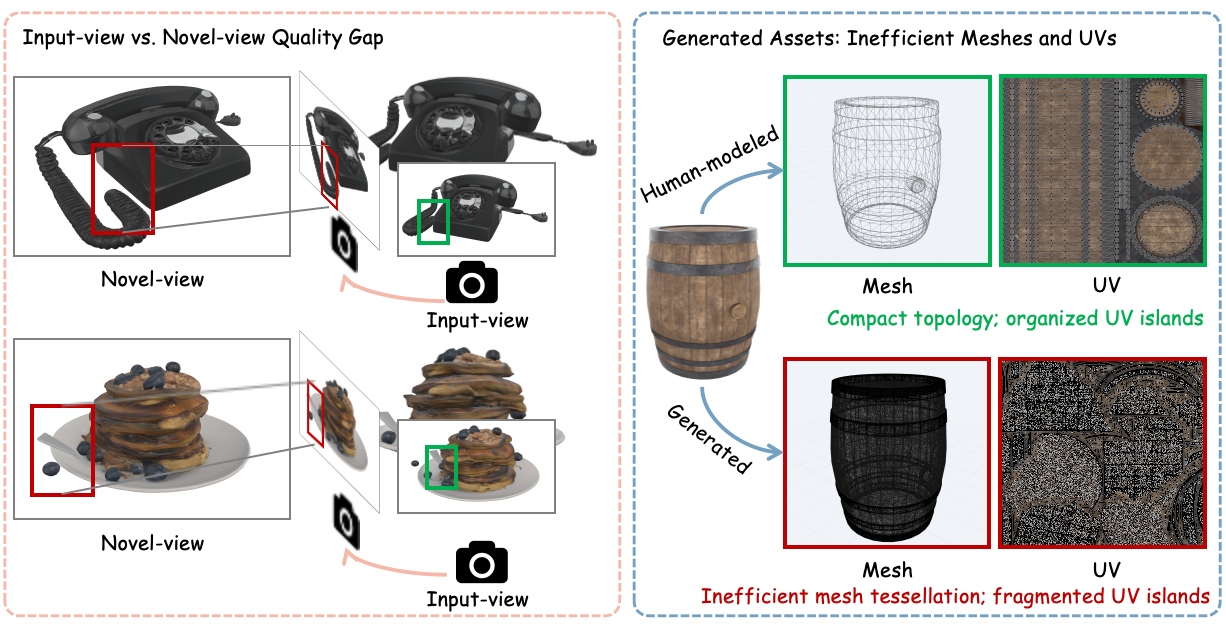}
\caption{Representative failure cases motivating Cyc3D. Left: an image-to-3D result may look plausible in the input view but reveal missing structure, distorted details, or unstable appearance from novel viewpoints, motivating cross-view object consistency. Right: generated assets can also suffer from native representation defects, such as inefficient mesh tessellation and fragmented UV islands, even when their rendered appearance is acceptable. These cases show why Cyc3D evaluates both viewpoint-dependent object consistency and representation-level asset quality.}
\label{fig:failure_modes}
\end{figure*}

We ground Cyc3D in recurring failure modes of image-to-3D generation, illustrated in Fig.~\ref{fig:failure_modes}. \emph{Reference drift} occurs when the generated object loses identity cues, proportions, colors, textures, or local details specified by the input image. \emph{Structural instability} occurs when the object changes identity, deforms globally, loses parts, duplicates structures, or collapses in detail under viewpoint-conditioned regeneration. \emph{Representation-level defects} occur when the explicit asset contains poor mesh discretization, invalid topology, distorted UV mapping, missing UV coordinates, or inconsistent texture parameterization.

These failures point to two reliability requirements. First, the model should form a \emph{stable 3D object hypothesis}: different observations of the same object should lead to compatible object interpretations rather than semantic drift or unstable part structures. Second, the model should realize this hypothesis as a \emph{faithful and valid 3D asset}: the output should preserve the conditioning image while providing geometry, surface parameterization, and texture maps that can be rendered, edited, converted, and reused. Cyc3D therefore organizes evaluation into \emph{Cross-View Object Consistency} and \emph{Representation Quality}; Sec.~\ref{sec:cyc3d_eval} instantiates these requirements as concrete protocols and metrics.

\section{Cyc3D Eval}
\label{sec:cyc3d_eval}

\subsection{Evaluation Overview}

\begin{figure*}[t]
\centering
\includegraphics[width=\textwidth]{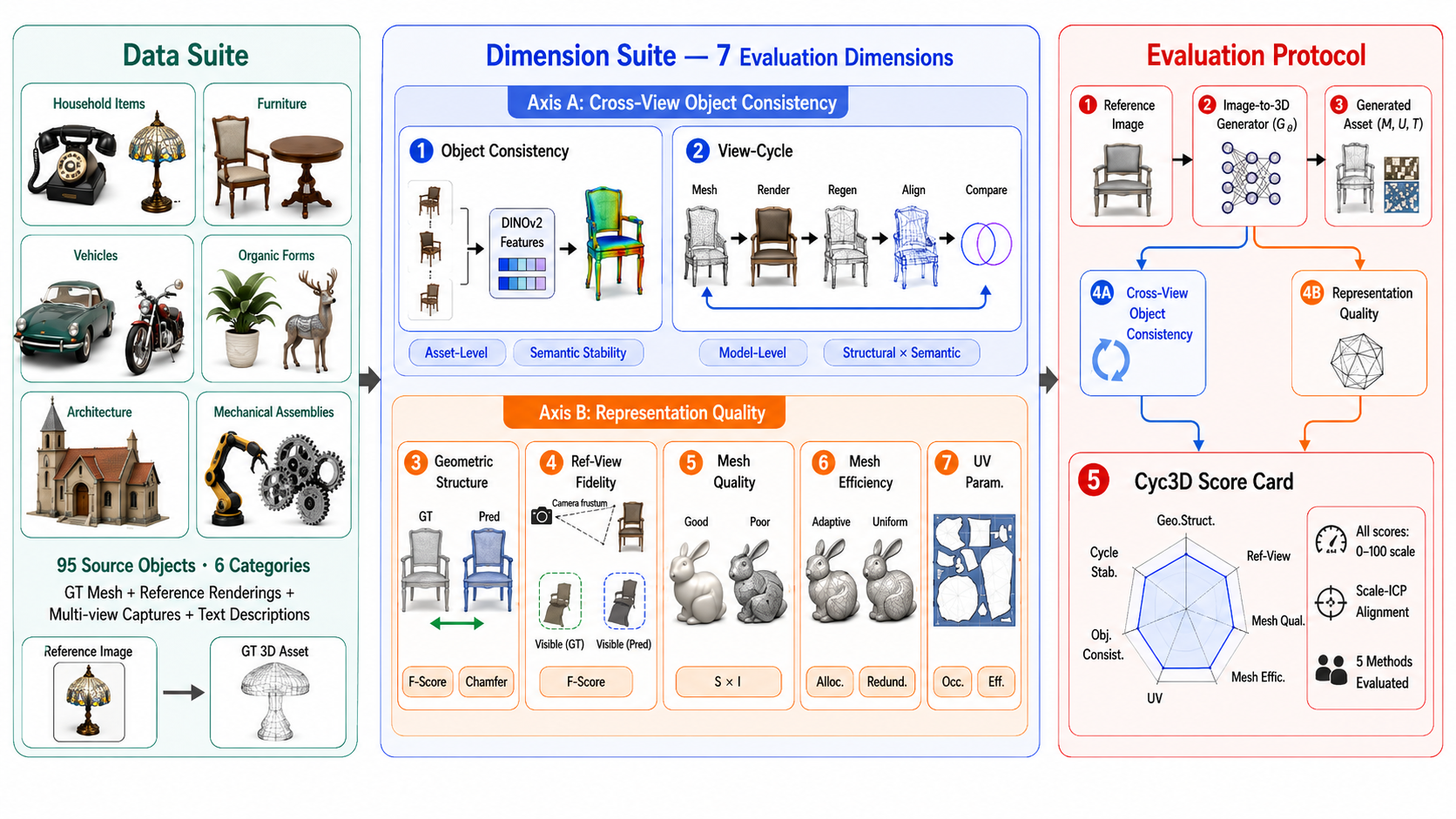}
\caption{Overview of the Cyc3D evaluation framework. Starting from a reference image, an image-to-3D generator produces a textured asset with geometry, UV parameterization, and texture. Cyc3D evaluates the generator at the model level through View-Cycle Structural Consistency, and evaluates the generated asset through reference-image fidelity, mesh discretization quality, UV parameterization quality, and multi-view object consistency.}
\label{fig:overview}
\end{figure*}

Our goal is to evaluate image-to-3D generation not only by how plausible the rendered views look, but by whether the generator has formed a reliable 3D understanding and produced a usable 3D asset. As shown in Fig.~\ref{fig:overview}, Cyc3D organizes evaluation along two complementary axes. \emph{Cross-View Object Consistency} asks whether the generated object maintains a coherent identity across viewpoints and whether the generator's 3D interpretation is stable under re-observation. \emph{Representation Quality} asks whether the output asset---its geometry, mesh structure, UV parameterization, and reference-view fidelity---meets the requirements of practical 3D workflows beyond single-view rendering.

\subsection{Cross-View Object Consistency}

A reliable 3D generator should produce objects whose identity and structure remain stable regardless of the viewing angle. Cyc3D tests this requirement at two levels: first, whether a single generated asset is internally consistent across viewpoints, and second, whether the generator itself maintains a stable 3D interpretation when repeatedly confronted with its own outputs from novel views.

\subsubsection{Asset-level Object Consistency}

A common failure of current generators is the \emph{Janus problem}: the asset looks plausible from the input view, but reveals incompatible content from other angles---multiple faces, contradictory textures, or semantically incoherent regions. The key insight is that if a 3D surface truly represents a single coherent object, then a vision foundation model should assign each surface point a consistent semantic identity regardless of the observing viewpoint.

We exploit this by rendering the generated mesh from multiple viewpoints, extracting dense feature maps from a self-supervised vision model, and projecting the pixel-level features back onto mesh vertices through visibility-aware reprojection. For each vertex visible in multiple views, we measure the variance of its associated features across those views. Vertices with high cross-view variance indicate semantic inconsistency. The final score is the fraction of surface vertices that remain semantically stable across all observing viewpoints---a simple yet effective proxy for whether the generator has formed a unified object interpretation rather than pasting view-dependent appearances onto the surface.

\subsubsection{View-Cycle Structural Consistency}

Asset-level consistency only checks a single fixed output. A stronger test asks: does the generator form a \emph{stable 3D prior}? If we render a generated asset from a novel viewpoint and feed that rendering back into the same generator, it should reproduce a structurally compatible object rather than drifting to a different interpretation. Repeating this process creates a closed-loop stress test for the model's underlying 3D understanding.

Concretely, given an initial asset $M_0$ and a cyclic camera sequence $\{c_i\}$, Cyc3D iterates a \emph{render--regenerate--align} loop: at each step, the current mesh is rendered from a novel viewpoint, passed through the generator to produce a new mesh, and rigidly aligned back to the previous step before comparison (Fig.~\ref{fig:view-cycle}). The per-step score combines two complementary signals:

\begin{figure*}[t]
\centering
\includegraphics[width=\textwidth]{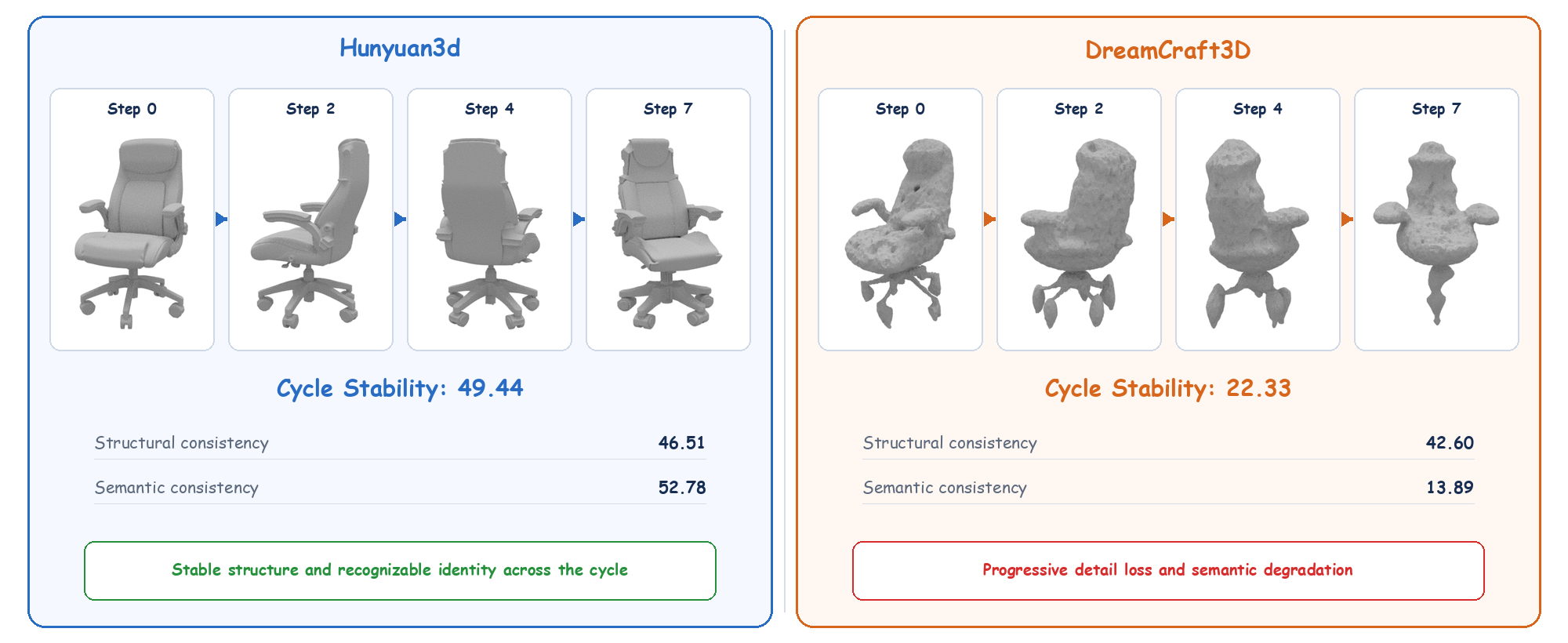}
\caption{Evaluation examples from Cyc3D. While both generations resemble an office chair at the beginning of the cycle, Hunyuan3d preserves the chair's structure and identity across viewpoints. In contrast, DreamCraft3D progressively loses the chair base, armrests, and backrest details (reflected in structural consistency), and eventually becomes less recognizable as an office chair (reflected in semantic consistency).}
\label{fig:view-cycle}
\end{figure*}
\begin{equation}
\mathrm{Cycle}_i = \sqrt{\mathrm{Str}_i \cdot \mathrm{Sem}_i},
\end{equation}
where $\mathrm{Str}_i$ aggregates geometric overlap (F-Score) and perceptual similarity (SSIM, LPIPS) between the previous and regenerated meshes within the visible frustum, while $\mathrm{Sem}_i$ captures VLM-assessed semantic recognizability anchored by the object's description. The geometric mean enforces an AND relation: a model cannot score highly by preserving geometry while losing identity, or by remaining recognizable while its shape drifts. The final score averages $\mathrm{Cycle}_i$ over all steps. In addition, Cyc3D reports the weakest step and the per-step trend slope, distinguishing gradual degradation from sudden collapse.

\subsection{Representation Quality}

A generated 3D asset is not merely an image-like rendering---it is a structured representation that should be rendered, edited, textured, converted, and reused. Many existing benchmarks evaluate generated assets through rendered views alone, which can mask representation-level defects. Cyc3D complements perceptual evaluation with explicit analysis of the asset's geometric fidelity, mesh structure, and UV parameterization.

\subsubsection{Geometric Structure}

The most fundamental question about a generated object is whether its shape is correct. We measure this by aligning the predicted mesh to the ground-truth coordinate frame and computing bidirectional surface distances together with thresholded overlap (F-Score) over dense point-cloud samples. This yields a holistic score reflecting how faithfully the generated geometry reproduces the target's overall 3D structure, independent of texture or viewpoint.

\subsubsection{Reference-image Fidelity}

Image-to-3D generation differs from text-to-3D in a crucial way: the input image is not merely a semantic prompt, but an instance-level visual constraint. A generated asset should reproduce not just the correct category, but the specific proportions, details, and local geometry visible in the conditioning image. We capture this by restricting the geometric comparison to the visible region within the reference camera frustum. The resulting score specifically penalizes cases where the generator ``hallucinated'' unseen regions correctly but failed to preserve what the user explicitly provided.

\subsubsection{Mesh Quality and Efficiency}

A mesh may achieve high geometric fidelity to the target shape yet still be unsuitable for practical use---its surface may be noisy from marching-cubes extraction, or it may have collapsed into a degenerate blob. Conversely, a clean mesh may waste its face budget by distributing triangles uniformly rather than concentrating them where geometric complexity demands. Cyc3D therefore evaluates both surface quality and tessellation efficiency.

\emph{Mesh Quality} captures whether the surface is well-formed. The intuition is that a high-quality mesh should be simultaneously smooth (free of high-frequency noise) and structurally intact (not collapsed). We decompose it as:
\begin{equation}
\mathrm{MeshQuality} = \mathrm{Smoothness} \times \mathrm{Integrity}.
\end{equation}
For Smoothness, we measure the high-frequency component of normal deviation between the predicted surface and the ground truth: after computing per-point angular errors, we extract each point's deviation from its local spatial median, isolating noise from legitimate geometric variation. The resulting residual is mapped to a score via per-object calibration using GT-derived clean and noisy reference meshes, avoiding reliance on absolute thresholds. For Integrity, we detect collapse by measuring (1) the fraction of predicted points that fall within a loose neighborhood of the GT surface, and (2) the asymmetry between forward and reverse distance quantiles---a collapsed blob produces few GT-aligned points and highly asymmetric distances.

\emph{Mesh Efficiency} asks a complementary question: given that a mesh is well-formed, does it allocate faces where they matter? The key idea is to factor out raw face count by simplifying both the prediction and the GT reference to a common budget via quadric-error decimation, then comparing \emph{where} faces are placed rather than how many exist. We voxelize the normalized meshes into a spatial grid and compute face-density histograms. The allocation score is the cosine similarity between the predicted and ground-truth histograms:
\begin{equation}
A = \frac{\mathbf{h}_{\mathrm{pred}} \cdot \mathbf{h}_{\mathrm{gt}}}{\|\mathbf{h}_{\mathrm{pred}}\| \cdot \|\mathbf{h}_{\mathrm{gt}}\|}.
\end{equation}
Importantly, the complexity reference comes exclusively from the GT mesh, so noisy surfaces cannot inflate their score by ``needing more faces.''

\subsubsection{UV Parameterization Quality}

Even when geometry and texture look good in rendered views, the underlying UV layout may be poorly suited for editing or re-texturing. Large blank regions waste texture resolution, excessive island fragmentation introduces padding overhead, and long seams risk visible discontinuities at render time. These defects are invisible in a single rendered frame but directly impact asset usability.

We evaluate UV quality through three complementary signals: \emph{occupancy} (the fraction of texture pixels actually used), \emph{effective utilization} (occupancy discounted by padding overhead from island fragmentation), and \emph{seam compactness} (the fraction of mesh edges that do not introduce UV discontinuities). The final UV score aggregates these signals into a single measure:
\begin{equation}
\mathrm{UVScore}=
\Phi_{\mathrm{uv}}\left(R_{\mathrm{uv}},R_{\mathrm{eff}},1-R_{\mathrm{seam}}\right).
\end{equation}
A high score indicates a UV layout that covers texture space densely, remains compact after island padding, and minimizes seam artifacts---properties that determine whether the asset can be reliably re-textured, edited, or converted to other formats.

\section{Experiments}
\label{sec:experiments}

We first describe the evaluation dataset and target methods (Sec.~\ref{sec:dataset}--\ref{sec:methods}), then present implementation details (Sec.~\ref{sec:impl}), and finally discuss findings from the main results (Sec.~\ref{sec:results}).

\subsection{Dataset}
\label{sec:dataset}

Cyc3D constructs an evaluation suite of 95 source 3D objects spanning household items, furniture, vehicles, organic forms, architectural structures, and mechanical assemblies. Each object is paired with a ground-truth mesh, canonical reference-view renderings, multi-view renderings for consistency evaluation, and optional text descriptions for VLM-assisted scoring. The dataset is designed to cover a broad spectrum of geometric complexity---from simple convex objects to intricate structures with thin parts, holes, and fine details---so that the benchmark can distinguish generators that handle only easy cases from those with robust 3D priors.

\subsection{Evaluated Methods}
\label{sec:methods}

We evaluate five representative image-to-3D methods that span the current landscape:

\begin{itemize}
    \item \textbf{Hunyuan3D} and \textbf{Tripo3D v3.1}: closed-source feed-forward systems that directly produce textured meshes from a single image, representing the strongest available commercial generators.
    \item \textbf{Stable Zero123}, \textbf{Magic123}, and \textbf{DreamCraft3D}: open-source optimization-based pipelines that lift images to 3D via score distillation or multi-view synthesis, representing accessible research baselines with varying pipeline complexity.
\end{itemize}

\subsection{Implementation Details}
\label{sec:impl}

All metrics are computed on a unified 0--100 scale (higher is better). Metrics requiring ground-truth reference align the predicted mesh to the GT coordinate frame via Scale-ICP before scoring. Mesh Efficiency uses a unified face budget via QEM simplification for cross-method comparability. Cycle Stability runs multiple render--regenerate steps per object along a horizontal-orbit camera sequence with uniformly spaced azimuth increments; each step samples surface points within the visible frustum for geometric comparison. The semantic signal is provided by a vision-language model rating object recognizability on an absolute scale, anchored by metadata descriptions. Object Consistency renders multiple views per object and extracts self-supervised vision features (DINOv2) upsampled to pixel resolution, following the vertex-level projection and consistency protocol of Eval3D~\cite{duggal2025eval3d}.

\subsection{Main Results}
\label{sec:results}

Table~\ref{tab:main_results} summarizes the quantitative results across all evaluation dimensions.

\begin{table*}[t]
\centering
\small
\setlength{\tabcolsep}{4.5pt}
\renewcommand{\arraystretch}{1.2}
\caption{Cyc3D results of image-to-3D models. We benchmark the five methods discussed in Sec.~\ref{sec:methods} on all evaluation dimensions. For each metric, the score ranges from 0 to 100. Higher values indicate better performance.}
\label{tab:main_results}
\resizebox{\linewidth}{!}{
\begin{tabular}{llccccccc}
\toprule
& & \multicolumn{5}{c}{\textbf{Representation Quality}}
& \multicolumn{2}{c}{\textbf{Cross-View Object Consistency}} \\
\cmidrule(lr){3-7} \cmidrule(lr){8-9}
Method & Type & Geo. Struct.\,$\uparrow$ & Ref-View Fid.\,$\uparrow$ & Mesh Qual.\,$\uparrow$ & Mesh Effic.\,$\uparrow$ & UV Param.\,$\uparrow$ & Object Consist.\,$\uparrow$ & Cycle Stab.\,$\uparrow$ \\
\midrule
Hunyuan3D & Closed & \cellb{82.6} & \cellb{70.2} & \cellu{83.9} & \cellb{89.1} & \cellu{68.8} & \cellu{83.7} & \cellb{47.1} \\
Tripo3D v3.1 & Closed & \cellu{75.2} & \cellu{59.1} & \cellb{84.2} & \cellu{86.9} & \cellb{78.8} & \cellb{87.3} & \cellu{45.6} \\
\midrule
Stable Zero123 & Open & \cell{45.3} & \cell{34.5} & \cell{41.6} & \cell{51.5} & \cell{70.0} & \cell{54.9} & \cell{25.5} \\
Magic123 & Open & \cell{51.0} & \cell{37.4} & \cell{59.3} & \cell{62.1} & \cell{70.5} & \cell{62.6} & \cell{27.6} \\
DreamCraft3D & Open & \cell{63.1} & \cell{42.4} & \cell{33.8} & \cell{75.2} & \cell{69.5} & \cell{72.1} & \cell{27.8} \\
\bottomrule
\end{tabular}}
\end{table*}

Hunyuan3D and Tripo3D v3.1 rank first or second on nearly all dimensions and produce smoother, more complete surfaces than the optimization-based baselines. This gap is consistent with feed-forward systems learning stronger category-level 3D priors from large-scale training, whereas per-instance optimization relies more heavily on 2D supervision. Because the commercial training data and objectives are unavailable, however, we cannot disentangle the effects of data, architecture, and supervision.

The cross-metric results reveal failures that a single quality score would obscure. DreamCraft3D has the lowest Mesh Quality ($33.8$) but competitive Mesh Efficiency ($75.2$): its faces are allocated to complex regions, yet the extracted surface remains noisy. Likewise, Stable Zero123's low Object Consistency ($54.9$) corresponds to repeated faces, contradictory textures, and incoherent back views. These observations suggest that plausible per-view optimization does not necessarily enforce a clean or globally coherent 3D asset.

Most importantly, high single-pass quality does not imply a stable 3D prior. Even Hunyuan3D and Tripo3D v3.1 achieve only $47.1$ and $45.6$ in Cycle Stability, while all optimization-based methods remain below $28$. As illustrated in Fig.~\ref{fig:view-cycle}, reconstruction errors accumulate when each output becomes the input to the next step. A likely reason is that current generators are trained for one-step reconstruction rather than cycle invariance, while imperfect novel-view renderings introduce a distribution shift that compounds across iterations. Stronger priors delay this drift but do not eliminate it.

\section{Related Work}

\paragraph{3D Generation.}
Recent 3D generation methods have evolved from per-instance optimization driven by 2D generative priors toward feed-forward reconstruction and explicit asset generation. Early text-to-3D systems commonly distill pretrained text-to-image diffusion models into NeRF, SDF, or hybrid 3D representations, enabling open-vocabulary generation without large-scale paired text--3D supervision. Magic3D accelerates this process with a coarse-to-fine pipeline that refines a coarse representation into a textured mesh, while ProlificDreamer improves score distillation through variational score distillation, producing more detailed and diverse 3D content~\cite{lin2023magic3d,wang2023prolificdreamer}. Latent-NeRF shows that text-only optimization is under-constrained and benefits from explicit shape guidance, and Fantasia3D disentangles geometry and appearance to improve geometry learning, material modeling, and compatibility with downstream graphics pipelines~\cite{metzer2023latent,chen2023fantasia3d}.

In parallel, a growing body of work has moved closer to directly producing usable 3D assets. GET3D learns to generate explicit textured meshes with rich geometry and high-fidelity textures from image collections~\cite{gao2022get3d}. Image-conditioned methods such as One-2-3-45 and One-2-3-45++ convert a single reference image into a full 360-degree textured mesh by combining multi-view generation with 3D reconstruction or 3D diffusion~\cite{liu2023one,liu2024one}. Large reconstruction and mesh-native models further emphasize feed-forward inference, explicit 3D inductive bias, and high-quality mesh extraction~\cite{hong2024lrm,liu2024meshformer}. These advances indicate that modern 3D generation is no longer concerned only with whether a rendered view looks plausible. Practical systems are increasingly expected to preserve the conditioning image, maintain stable 3D structure, and output assets that are usable in standard graphics pipelines. This trend motivates evaluation protocols that go beyond rendered appearance and prompt alignment, and explicitly measure reference-image fidelity, mesh quality, UV quality, and structural stability.

\paragraph{Benchmarks for 3D Generation.}
Evaluation for 3D generation has recently shifted from single proxy scores toward human-aligned, multidimensional, and interpretable protocols. GPTEval3D introduces a scalable preference-based paradigm in which GPT-4V compares rendered multi-view assets under user-defined criteria and aggregates pairwise judgments with Elo ratings~\cite{wu2024gpt}. Gen3DEval follows the same need for scalable human-aligned evaluation, but trains a vision-language model for pairwise assessment of generated 3D objects along text fidelity, appearance, and surface quality~\cite{maiti2025gen3deval}. MATE-3D further emphasizes subjective multidimensional quality assessment by collecting human scores over semantic alignment, geometry quality, texture quality, and overall quality across diverse prompt categories~\cite{zhang2025benchmarking}. More recent benchmarks also make evaluation more diagnostic: Eval3D decomposes 3D quality into geometric, structural, semantic, text-alignment, and aesthetic aspects using interpretable probes, while Hi3DEval extends evaluation hierarchically from object-level quality to part-level and material-level validity~\cite{duggal2025eval3d,zhang2025hi3deval}.

These benchmarks provide important foundations for asset-level assessment, where a generated 3D asset is evaluated from rendered views, captions, normal maps, or part and material annotations. Cyc3D builds on this line of work by making two image-to-3D-specific requirements explicit: the conditioning image should be treated as an instance-level visual constraint, and the generated output should be evaluated as a native 3D asset rather than only through rendered-view proxies. These requirements motivate our focus on reference-view fidelity, mesh and UV quality, and model-level cycle stability.

\section{Conclusion}

We have presented Cyc3D, a comprehensive evaluation framework for image-to-3D generation that assesses both cross-view object consistency and representation quality. By introducing View-Cycle Structural Consistency as a model-level metric and incorporating explicit mesh and UV quality evaluation alongside reference-image fidelity, Cyc3D provides a more complete picture of image-to-3D reliability than existing protocols that focus primarily on rendered appearance. Our evaluation of five representative methods reveals that while recent closed-source systems produce high-quality assets with reasonable geometric fidelity, all current generators still fail to maintain stable 3D object hypotheses under cyclic regeneration. This finding highlights the gap between visually plausible generation and robust 3D object understanding, and motivates future work on improving the structural consistency of image-conditioned 3D generators.

\bibliography{aaai2027}

\end{document}